# Local Feature or Mel Frequency Cepstral Coefficients - Which One is Better for MLN-Based Bangla Speech Recognition?

Foyzul Hassan[1], Mohammed Rokibul Alam Kotwal[1], Md. Mostafizur Rahman[1], Mohammad Nasiruddin[2], Md. Abdul Latif[2] and Mohammad Nurul Huda[1]

[1] United International University
Dhaka, Bangladesh

[2] University of Asia Pacific
Dhaka, Bangladesh
foyzul.hassan@gmail.com, rokib_kotwal@yahoo.com, tuhin_cse@yahoo.com,
mohammadnasiruddin@gmail.com, csefusion@yahoo.com, mnh@cse.uiu.ac.bd

**Abstract.** This paper discusses the dominancy of local features (LFs), as input to the multilayer neural network (MLN), extracted from a Bangla input speech over mel frequency cepstral coefficients (MFCCs). Here, LF-based method comprises three stages: (i) LF extraction from input speech, (ii) phoneme probabilities extraction using MLN from LF and (iii) the hidden Markov model (HMM) based classifier to obtain more accurate phoneme strings. In the experiments on Bangla speech corpus prepared by us, it is observed that the LF-based automatic speech recognition (ASR) system provides higher phoneme correct rate than the MFCC-based system. Moreover, the proposed system requires fewer mixture components in the HMMs.

## 1 Introduction

Various methods have been proposed for obtaining an accurate automatic speech recognition (ASR) system [1-6]. Although some of them perform adequately, most hidden Markov model (HMM) based methods have several limitations. For example, (a) they require a large number of speech parameters and a large speech corpus to solve coarticulation problems using context-sensitive triphone models, and (b) they require a higher computational cost to achieve an acceptable performance.

On the other hand, many neural network based methods are approached to reduce speech parameters [7-10] and to achieve higher recognition performance. Some of these methods solve coarticulation problems by embedding context dependent input vectors. Most of these methods use mel frequency cepstral coefficients (MFCCs) parameters of speech signal as input vector to multilayer neural network (MLN) [7-10] and provides comparable recognition performance using higher computation cost.

In this paper, we proposed a Bangla phoneme recognition system for an ASR by inputting local features (LFs) instead of MFCCs. The method consists of three stages: (i) the first stage extracts LFs from the Bangla input speech, (ii) the second stage incorporates an MLN to obtain phoneme probabilities for the LFs extracted at first stage and (iii) the final stage embeds an HMM-based classifier to output phoneme strings. This study shows that the LF-based system provides higher phoneme correct rate (PCR) than the system based on MFCCs. For designing the proposed and existing methods, we have prepared a medium size Bangla speech corpus and done the following experiments: (i) MFCC39+MLN+HMM and (ii) LF25+MLN+HMM [Proposed].

The paper is organized as follows: Section II shows the preparation of Bangla speech corpus. Section III explains the Bangla phonemes and corresponding IPA, while the LF extraction procedure is described in Section IV. The system configuration of the existing phoneme recognition methods with the proposed method is discussed in the Section V. Experimental setup are provided in Section VI, while experimental results are analyzed in Section VII. Finally, Section VIII draws some conclusion.

## 2   Bangla Speech Corpus

At present, a real problem to do experiment on Bangla phoneme ASR is the lack of proper Bangla speech corpus. In fact, such a corpus is not available or at least not referenced in any of the existing literature. Therefore, we develop a medium size Bangla speech corpus, which is described below.

Hundred sentences from the Bengali newspaper "Prothom Alo" [11] are uttered by 30 male speakers of different regions of Bangladesh. These sentences (30x100) are used for training corpus (D1). On the other hand, different 100 sentences from the same newspaper uttered by 10 different male speakers (total 1000 sentences) are used as test corpus (D2). All of the speakers are Bangladeshi nationals and native speakers of Bangla. The age of the speakers ranges from 20 to 40 years. We have chosen the speakers from a wide area of Bangladesh: Dhaka (central region), Comilla – Noakhali (East region), Rajshahi (West region), Dinajpur – Rangpur (North-West region), Khulna (South-West region), Mymensingh and Sylhet (North-East region). Though all of them speak in standard Bangla, they are not free from their regional accent.

Recording was done in a quiet room located at United International University (UIU), Dhaka, Bangladesh. A desktop was used to record the voices using a head mounted close-talking microphone. We record the voice in a place, where ceiling fan and air conditioner were switched on and some low level street or corridor noise could be heard.

Jet Audio 7.1.1.3101 software was used to record the voices. The speech was sampled at 16 kHz and quantized to 16 bit stereo coding without any compression and no filter is used on the recorded voice.

## 3   Bangla Phonemes Schemes

### 3.1   Bangla Phonemes

The Phonetic inventory of Bangla consists of 14 vowels, including seven nasalized vowels, and 29 consonants. An approximate phonetic scheme in IPA is given in Table 1. In Table 1 (a), only the main 7 vowel sounds are shown, though there exists two more long counterpart of /i/ and /u/, denoted as /i:/ and /u:/, respectively. These two long vowels are seldom pronounced differently than their short counterparts in modern Bangla. There is controversy on the number of Bangla consonants.

Native Bangla words do not allow initial consonant clusters: the maximum syllable structure is CVC (i.e. one vowel flanked by a consonant on each side) [12]. Sanskrit words borrowed into Bangla possess a wide range of clusters, expanding the maximum syllable structure to CCCVC. English or other foreign borrowings add even more cluster types into the Bangla inventory.

**Table 1.**   Bangla phonetic scheme in IPA extracted from http://en.wikipedia.org/wiki/Bengali_phonology.

(a) Vowel

|  | Front | Central | Back |
|---|---|---|---|
| Close | i<br>i |  | u<br>u |
| Close-mid | e<br>e |  | o<br>o |
| Open-mid | æ<br>ê |  | ɔ<br>ô |
| Open |  | a<br>a |  |

(b) Consonants

|  |  | Labial | Dental/<br>Alveolar | Retroflex | Lamino-<br>Postalveolar | Velar | Glottal |
|---|---|---|---|---|---|---|---|
| Nasal |  | m<br>m | n<br>n |  |  | ŋ<br>ng |  |
| Plosive | voiceless | p<br>p | t̪<br>t | ʈ<br>ṭ | tʃ<br>ch | k<br>k |  |
|  | aspirated | pʰ[1]<br>ph | t̪ʰ<br>th | ʈʰ<br>ṭh | tʃʰ[2]<br>chh | kʰ<br>kh |  |
|  | voiced | b<br>b | d̪<br>d | ɖ<br>ḍ | dʒ<br>j | g<br>g |  |
|  | murmured | bʰ<br>bh | d̪ʰ<br>dh | ɖʰ<br>ḍh | dʒʰ[3]<br>jh | gʰ<br>gh |  |
| Fricative |  | f[1]<br>f | s², z³<br>s |  | ʃ[2]<br>sh |  | h<br>h |
| Approximant |  |  | l<br>l |  |  |  |  |
| Rhotic |  |  | r[4]<br>r | ɽ[4]<br>ṛ |  |  |  |

### 3.2   Bangla Words

Table 2 lists some Bangla words with their written forms and the corresponding IPA. From the table, it is shown that the same 'আ' (/a/) has different pronunciation based on succeeding phonemes 'ম', 'চ' and 'ব'. These pronunciations are sometimes long or short. For long and short 'আ' we have used two different phonemes /aa/ and /ax/, respectively. Similarly, we have considered all variations of same phonemes and consequently, found total 51 phonemes excluding beginning and end silence (/sil/) and short pause (/sp/).

**Table 2.** Some Bangla words with their orthographic transcriptions and IPA.

| Bangla Word | English Pronunciation | IPA | Our Symbol |
|---|---|---|---|
| আমরা | AAMRA | /a m r a/ | /aa m r ax/ |
| আচরণ | AACHORON | /a tʃ r n/ | /aa ch ow r aa n/ |
| আবেদন | ABEDON | /a b æ d̪ n/ | /ax b ae d aa n/ |

## 4 Local Feature Extraction

At the acoustic feature extraction stage, the input speech is first converted into LFs that represent a variation in spectrum along the time and frequency axes. Two LFs, which are shown in Fig. 1, are then extracted by applying three-point linear regression (LR) along the time (t) and frequency (f) axes on a time spectrum pattern (TS), respectively. Fig. 2 exhibits an example of LFs for an input utterance. After compressing these two LFs with 24 dimensions into LFs with 12 dimensions using discrete cosine transform (DCT), a 25-dimensional (12 $\Delta t$, 12 $\Delta f$, and $\Delta P$, where P stands for the log power of a raw speech signal) feature vector called LF is extracted.

## 5 System Configuration

### 5.1 MFCC-based System

Fig. 3 shows the phoneme recognition method using MLN [13]. At the acoustic feature extraction stage, input speech is converted into MFCCs of 39 dimensions (12-MFCC, 12-$\Delta$MFCC, 12-$\Delta\Delta$MFCC, P, $\Delta P$ and $\Delta\Delta P$, where P stands for raw energy of the input speech signal). MFCCs are input to an MLN with four layers, including 3 hidden layers, after combining preceding (t-3)-th and succeeding (t+3)-th frames with the current t-th frame. The MLN has 53 output units (total 53 monophones) of phoneme probabilities for the current frame t. The three hidden layers consist of 400, 200 and 100 units, respectively. The MLN is trained by using the standard back-propagation algorithm. This method yields comparable recognition performance.

### 5.2 Proposed System

Fig. 4 shows the phoneme recognition method using MLN. At the acoustic feature extraction stage, input speech is converted into LFs of 25 dimensions (12 $\Delta t$, 12 $\Delta f$, and ΔP, where P stands for the log power of a raw speech signal). LFs are input to an MLN with four layers, including 3 hidden layers, after combining preceding (t-3)-th and succeeding (t+3)-th frames with the current t-th frame. The MLN has 53 output units (total 53 monophones) of phoneme probabilities for the current frame t. The architecture and training procedure of MLN are same as Section 5.1.

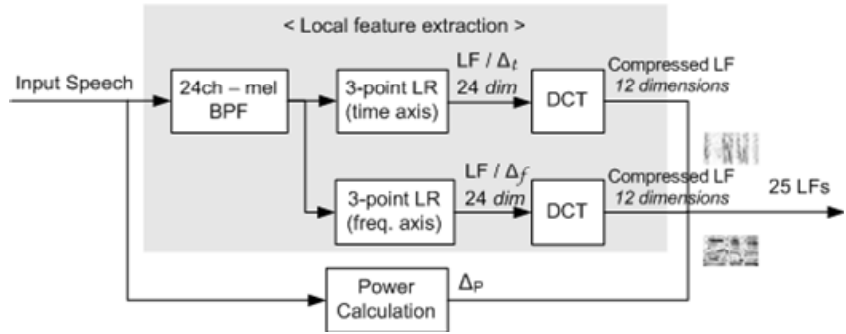

**Fig. 1.** LFs extraction procedure.

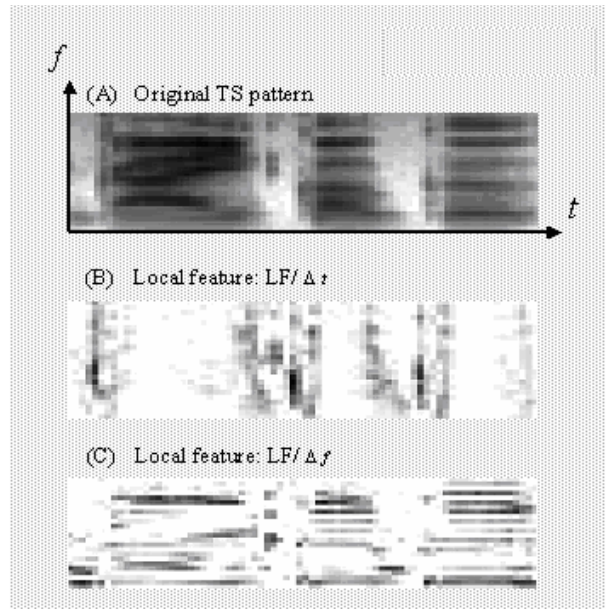

**Fig. 2.** Examples of LFs.

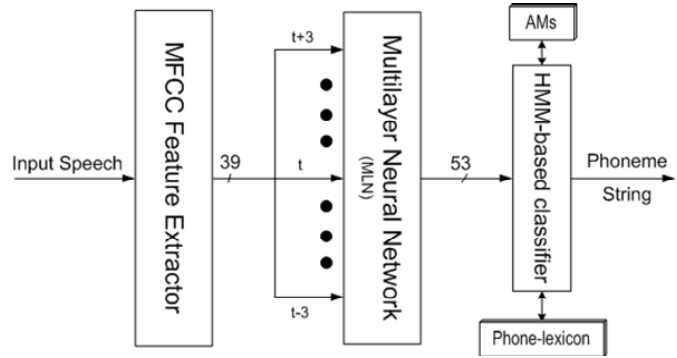

**Fig. 3.** MFCC-based Phoneme Recognition Method.

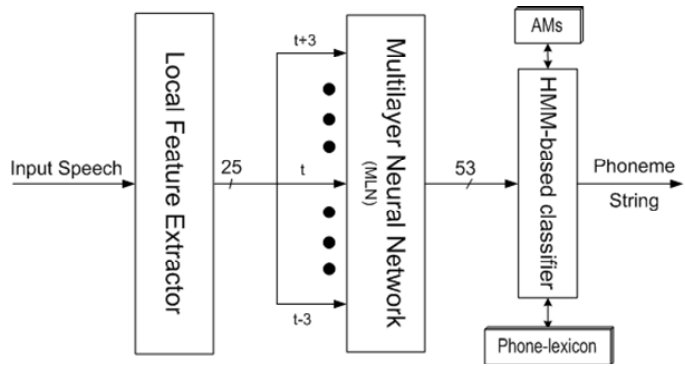

**Fig. 4.** LF-based Phoneme Recognition Method.

## 6   Experimental Setup

The frame length and frame rate are set to 25 ms and 10 ms (frame shift between two consecutive frames), respectively, to obtain acoustic features (MFCCs) from an input speech. MFCC comprised of 39 dimensional.

For designing an accurate phoneme recognizer, PCR for both D1 and D2 data set are evaluated using an HMM-based classifier. The D1 data set is used to design 53 Bangla monophones HMMs with five states, three loops, and left-to-right models. Input features for the HMM using both the methods is 53 dimensions. In the HMMs, the output probabilities are represented in the form of Gaussian mixtures, and diagonal matrices are used. The mixture components are set to 1, 2, 4, 8 and 16.

In our experiments of the MLN, the non-linear function is a sigmoid from 0 to 1 (1/(1+exp(-x))) for the hidden and output layers.

To evaluate PCR, we have prepared a medium size Bangla speech corpus and done the following experiments for both training (D1) and test (D2) data sets.

  (i)   MFCC39+MLN+HMM
  (ii)  LF25+MLN+HMM [Proposed]

## 7   Experimental Results and Discussion

Fig. 5 shows the comparison of PCR of training data set between MFCC39+MLN+HMM and LF25+MLN+HMM systems. It is observed from the figure that LF-based system always provides higher PCR than the conventional MFCC-based method. For an example, at mixture component 16, the LF-based system exhibits 61.07% phoneme correct rate, while 56.27% PCR is obtained by the MFCC-based method.

On the other hand, the PCR the test data (D2) are shown in the Fig. 6 for the investigated methods. The LF-based method outperformed the other methods for the evaluation of PCR. It is noted from mixture component 16 of Fig. 6 that the LF-based system having 55.02% correctness shows its better recognition performance over the MFCC-based method (52.07% PCR).

It is claimed that the proposed method reduces mixture components in HMMs and hence computation time. For an example from the Fig. 6, approximately 50.00% phoneme recognition correctness is obtained by the methods (i) and (ii) at mixture components eight and one, respectively.

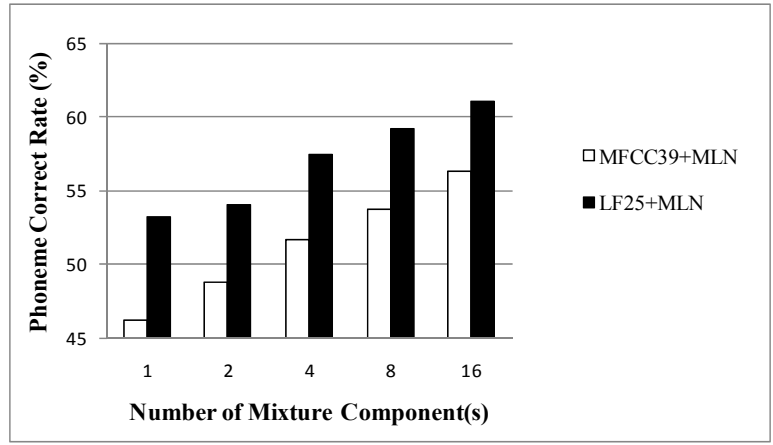

**Fig. 5.** Phoneme correct rate for training data set, D1.

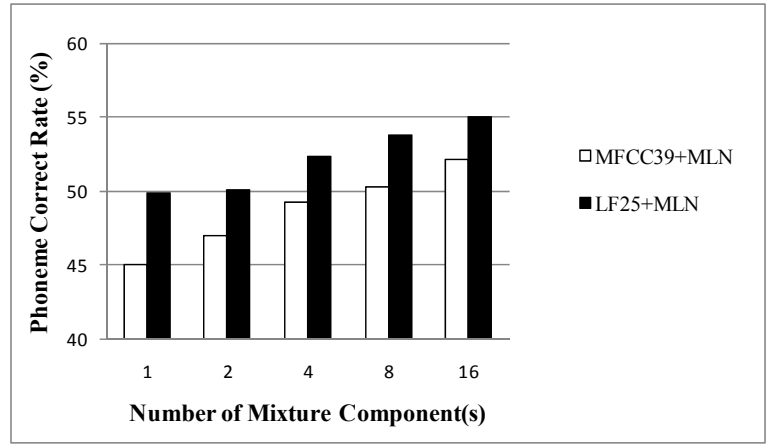

**Fig. 6.** Phoneme correct rate for test data set, D2.

## 6 Conclusion

In this paper, we proposed a Bangla phoneme recognition method for an ASR using local features. The following conclusions are drawn from the study.
(i) Our proposed method based on local features show higher phoneme correct rate for both training and test data set.
(ii) It requires fewer mixture components in HMMs.
(iii) Moreover, it reduces computation time because of lower dimensional (25 dim) LFs instead of higher dimensional MFCCs (39 dim).

In future, the author would like to do experiments using recurrent neural network (RNN). Moreover, Bangla word recognition using triphone model will be evaluated by the same proposed method in this paper.

## References

1. S. Nakagawa, et. al, "Noisy Speech Recognition Based on Integration/Selection of Multiple Noise Suppression Methods Using Noise GMMs," IEICE Trans. Inf & Syst., Vol. E91-D(3), pp.411-421, 2008
2. P. Jain, H. Hermansky and B. Kingsbury, "Distributed Speech Recognition Using Noise-Robust MFCC and TRAPS-Estimated Manner Features," Proc. ICSLP'02, Vol.I, pp.473-476, Sep. 2002
3. J. Marcus, et. al, " Phonetic recognition in a segment-based HMM," Proc. ICASSP, April 1993

4. P. Schwarz, et. al, " Hierarchical structures of neural networks for phoneme recognition," ICASSP, 2006
5. SUZUKI Hiroyuki, et. al,"Continuous Speech Recognition Based on General Factor Dependent Acoustic Models," IEICE transactions on information and systems Vol.E88-D, No.3(20050301) pp. 410-417, March 2005
6. S. Matsuda, T. Jitsuhiro, K. Markov, and S. Nakamura, "Speech recognition system robust to noise and speaking styles," Proc. ICSLP04, vol.IV, pp.2817-2820, 2004
7. K. Kirchhoff, et. al, "Combining acoustic and articulatory feature information for robust speech recognition," Speech Commun.,vol.37, pp.303-319, 2002
8. K. Kirchhoffs, " Robust Speech Recognition Using Articulatory information," Ph.D thesis, University of Bielefeld, Germany, July 1999
9. K. Roy, D. Das, and M. G. Ali, "Development of the speech recognition system using artificial neural network," in *Proc. 5$^{th}$ International Conference on Computer and Information Technology (ICCIT02)*, Dhaka, Bangladesh, 2002
10. M. R. Hassan, B. Nath, and M. A. Bhuiyan, "Bengali phoneme recognition: a new approach," in *Proc. 6$^{th}$ International Conference on Computer and Information Technology (ICCIT03)*, Dhaka, Bangladesh, 2003
11. Daily Prothom Alo. Online: www.prothom-alo.com
12. C. Masica, *The Indo-Aryan Languages*, Cambridge University Press, 1991
13. M. R. A. Kotwal "Neural Network Based Automatic Speech Recognition," M. Sc. Thesis 2010, United International University, Dhaka, Bangladesh